\documentclass[10pt, a4paper]{article}
\usepackage{lrec2022} 
\usepackage{multibib}
\newcites{languageresource}{Language Resources}
\usepackage{graphicx}
\usepackage{tabularx}
\usepackage{soul}
\usepackage{titlesec}
\titleformat{\section}{\normalfont\large\bfseries\center}{\thesection.}{1em}{}
\titleformat{\subsection}{\normalfont\SmallTitleFont\bfseries\raggedright}{\thesubsection.}{1em}{}
\titleformat{\subsubsection}{\normalfont\normalsize\bfseries\raggedright}{\thesubsubsection.}{1em}{}
\renewcommand\thesection{\arabic{section}}
\renewcommand\thesubsection{\thesection.\arabic{subsection}}
\renewcommand\thesubsubsection{\thesubsection.\arabic{subsubsection}}

\usepackage{epstopdf}
\usepackage[utf8]{inputenc}

\usepackage{hyperref}
\usepackage{xstring}

\usepackage{color}

\usepackage{microtype}
\usepackage[tbtags]{amsmath}
\usepackage{booktabs}
\usepackage{multirow}
\usepackage{xcolor}
\usepackage{colortbl}
\usepackage{standalone}
\usepackage{pgfplots}
\pgfplotsset{compat=newest}
\usepgfplotslibrary{colorbrewer}
\usepgfplotslibrary[groupplots,dateplot]
\usetikzlibrary[patterns,shapes.arrows]
\unexpanded{\def\startgroupplot{\groupplot}}
\unexpanded{\def\stopgroupplot{\endgroupplot}}

\usepackage{etoolbox}
\apptocmd{\sloppy}{\hbadness 10000\relax}{}{}

\definecolor{PaleYellow}{RGB}{255,230,204}
\definecolor{PaleRed}{RGB}{248,206,204}
\definecolor{PalePurple}{RGB}{225,213,231}
\definecolor{AquaBlue}{RGB}{176,227,230}
\newcommand\pale[1]{\cellcolor{PaleYellow}{#1}}
\newcommand{\hlfancy}[2]{\colorlet{hlcolor}{#1}\sethlcolor{hlcolor}\hl{#2}}

\title{Cyberbullying Classifiers are Sensitive to Model-Agnostic Perturbations}

\name{Chris Emmery$^{\includegraphics[width=8px]{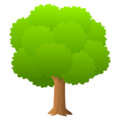},
                     \includegraphics[width=8px]{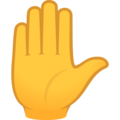}}$, 
\'{A}kos K\'{a}d\'{a}r$^{\includegraphics[width=8px]{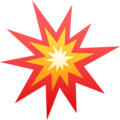}}$,
Grzegorz Chrupa\l{}a$^{\includegraphics[width=8px]{./gfx/tree.png}}$, 
Walter Daelemans$^{\includegraphics[width=8px]{./gfx/hand.png}}$}

\address{$^{\includegraphics[width=8px]{./gfx/tree.png}}$CSAI, Tilburg University, 
         $^{\includegraphics[width=8px]{./gfx/hand.png}}$CLiPS, University of Antwerp, 
         $^{\includegraphics[width=8px]{./gfx/explosion.png}}$Explosion \\
         \texttt{cmry@pm.me} \\}

\abstract{
A limited amount of studies investigates the role of model-agnostic adversarial behavior in toxic content classification. As toxicity classifiers predominantly rely on lexical cues, (deliberately) creative and evolving language-use can be detrimental to the utility of current corpora and state-of-the-art models when they are deployed for content moderation. The less training data is available, the more vulnerable models might become. This study is, to our knowledge, the first to investigate the effect of adversarial behavior and augmentation for cyberbullying detection. We demonstrate that model-agnostic lexical substitutions significantly hurt classifier performance. Moreover, when these perturbed samples are used for augmentation, we show models become robust against word-level perturbations at a slight trade-off in overall task performance. Augmentations proposed in prior work on toxicity prove to be less effective. Our results underline the need for such evaluations in online harm areas with small corpora. The perturbed data, models, and code are available for reproduction at \url{https://github.com/cmry/augtox}.
\\ \newline  }
\begin{document}

\maketitleabstract

\section{Introduction}

Our online presence has simplified contact with our (in)direct network, and thereby drastically changed how, and with whom we interact. While online connections and self-disclosure are often socially beneficial \cite{ValkePeter2007fw}, the absence of physical interaction has numerous adverse effects: it greatly reduces social accountability in (anonymous) interactions, amplifies one's exposure to people with malicious intent, and through our frequent use of mobile devices, the invasiveness thereof \cite{https://doi.org/10.1002/pits.20301}. These factors accumulate to persistent online toxic behavior---the scale of which online platforms continue to struggle with from a technical, legal, and ethical perspective.

Online harm \cite[provide a comprehensive taxonomy of this field]{banko-etal-2020-unified} and---particularly for Natural Language Processing (NLP)---abusive language, are highly complex phenomena. Their study spreads across several subfields (detection of hate speech, toxic comments, offensive and abusive language, aggression, and cyberbullying), all with their unique problem sets and (almost exclusively English) corpora \cite{10.1371/journal.pone.0243300}. Moreover, there are numerous open issues with these tasks, as highlighted in a range of critical studies \cite[for example]{DBLP:journals/corr/abs-1910-11922,ROSA2019333,swamy-etal-2019-studying,madukwe-etal-2020-data,DBLP:journals/corr/abs-2103-00153}. Those open issues primarily pertain to the contextual, historical, and multi-modal nature of toxicity, the specificity of the data, and poor generalization across domains.

\begin{figure}
\centering
\includegraphics[width=\columnwidth]{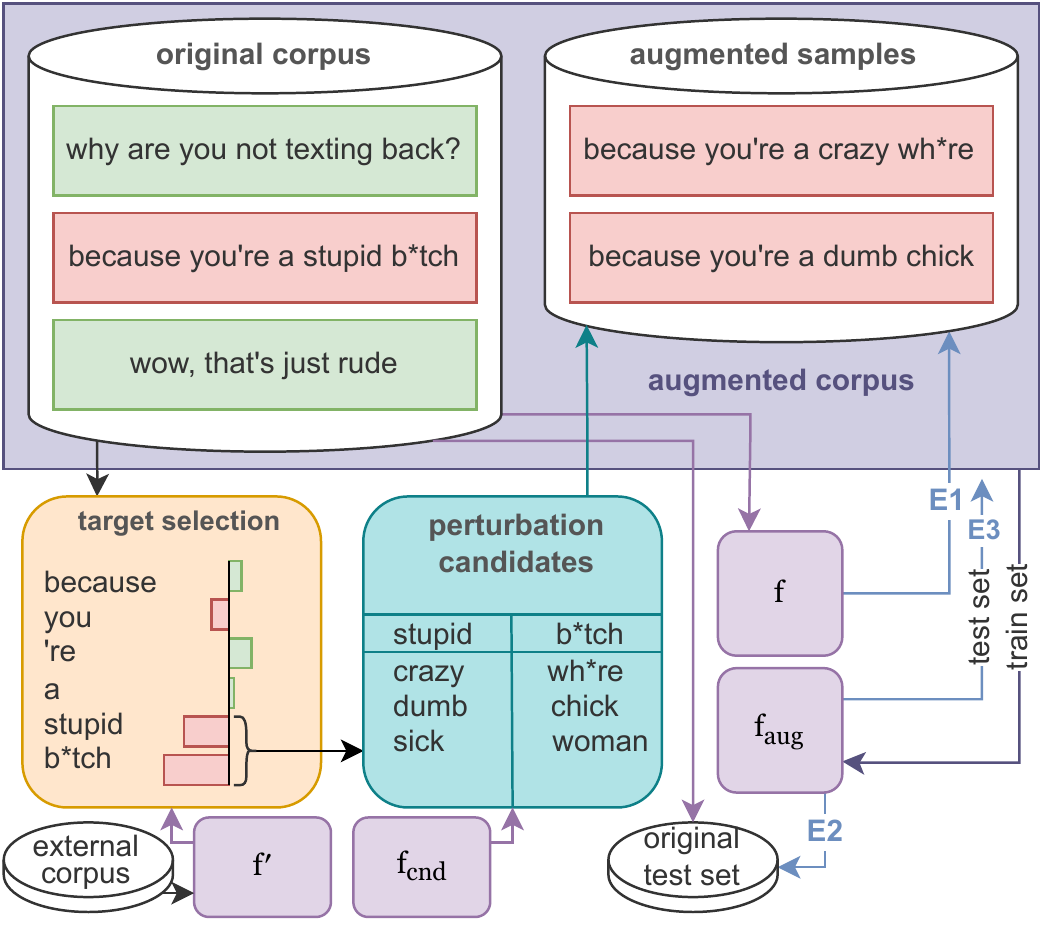}
\caption{Schematic overview of the presented experiments (E1-3) for data augmentation of cyberbullying content via model-agnostic lexical substitutions.}
\label{fig:paper}
\end{figure}

The current work focuses on one of these subproblems: the continuously evolving nature of \emph{toxic content}. Apart from the disparate channels and media through which (young) users communicate, this development particularly applies to the related vocabulary: slang, hate speech, or general insults (e.g., \emph{karen}, \emph{simp}, \emph{coofer}, and \emph{covidiot}). Given the strong focus on lexical cues exhibited by state-of-the-art toxic content classifiers \cite{gehman-etal-2020-realtoxicityprompts}, the existing corpora would have to be continuously expanded for models to retain their performance. This puts costly requirements on any system automatically moderating harmful content, while research in this domain still seems unconcerned with evaluating models in the wild \cite{DBLP:journals/corr/abs-2103-00153}.

The adversarial nature of toxicity exacerbates these issues further; similar to any security application, it is safe to assume malicious actors will try to (actively) subvert any form of moderation they are subjected to. Yet, while ample work has investigated systems toward mimicking such behavior \cite[for example, feature attacks against Google's Perspective API]{DBLP:journals/corr/HosseiniKZP17,ebrahimi-etal-2018-hotflip,DBLP:conf/ndss/LiJDLW19}, work on toxic content detection rarely incorporates tests for robustness against adversarial attacks. More importantly, these attacks are commonly tailored to an existing toxicity classifier, whereas a human adversary would not have direct access to the models performing moderation. A realistic implementation of subversive human behavior would therefore require model-agnostic attacks.

Accordingly, the current research combines multiple ideas from previous work: we apply lexical substitution to an online harm subtask with small corpora (cyberbullying detection) to investigate how lexical variation (either natural, or adversarial) affects model performance and, by extension, evaluate the robustness of current state-of-the-art models. We do this in a model-agnostic fashion; an external classifier indicates which words might be relevant to substitute. Those words are perturbed through a variety of transformer-based models, after which we assess changes in the predictions of a target classifier (see Figure~\ref{fig:paper}). The perturbations are \emph{not} selected to be adversarial against the external or target classifier. We subsequently evaluate to what extent augmenting existing cyberbullying corpora improves classifier performance, robustness against word-level perturbations, and transferability across different substitution models. With this, we provide new resources to evaluate, and compare, the robustness of future cyberbullying classifiers against lexical variation in toxicity.

\begin{table*}
    \centering
    \footnotesize
    \setlength{\tabcolsep}{5.5pt}
    \begin{tabular}{l|lll>{\columncolor{AquaBlue}}l>{\columncolor{AquaBlue}}ll>{\columncolor{AquaBlue}}llll>{\columncolor{AquaBlue}}lll}
        \toprule
         &   &       & &  \multicolumn{2}{c}{}          &     \multicolumn{3}{c}{\textsc{targets}}  \\
            \cmidrule{5-6} \cmidrule{8-8} \cmidrule{12-12}
        \textsc{prompt}  & You & are & a & \pale{r\hlfancy{black}{e}t\hlfancy{black}{a}rded} & \pale{dweeb}             & and & \pale{stupid}      & af & .        & Go & \pale{f\hlfancy{black}{u}ck}             & yourself  & . \\
        \textsc{tokens}  & You & are & a & \pale{r\hlfancy{black}{e}\ \#\#t\hlfancy{black}{a}r\ \#\#d\hlfancy{black}{e}d} & \pale{d\ \#\#we\ \#\#eb} & and & \pale{stupid}      & a\ \#\#f & .  & Go & \pale{f\hlfancy{black}{u}ck} & yourself & . \\
        \midrule
        \textsc{\#1} & You & are & a & silly     & baby                             & and & silly       & af & .        & Go & scr\hlfancy{black}{e}w             & yourself & .  \\
        \textsc{\#2} & You & are & a & useless   & teenager                         & and & dumb        & af & .        & Go & d\hlfancy{black}{i}ck             & yourself & .  \\
        \textsc{\#3} & You & are & a & sick      & b\hlfancy{black}{i}tch           & and & foolish     & af & .        & Go & sh\hlfancy{black}{i}t             & yourself &  .  \\
        \textsc{\#4} & You & are & a & crazy     & dog                              & and & useless     & af & .        & Go & d\hlfancy{black}{a}mn             & yourself &  .  \\
        \textsc{\#5} & You & are & a & dumb      & idiot                            & and & ignorant    & af & .        & Go & p\hlfancy{black}{i}ss             & yourself & .   \\
        \bottomrule
    \end{tabular}
    \caption{Lexical substitution example using Dropout BERT. Shows the (bowdlerized) initial \textsc{prompt}, which words are targeted for substitution (highlighted), their word-piece encoding (\textsc{tokens}), and the generated samples.}
    \label{tab:lex-sub}
\end{table*}

\section{Lexical Substitution}

We employ word-level or token-level perturbations (i.e., substitutions, see Table~\ref{tab:lex-sub} for examples), which implies that for a given target word $w_t$ in document $D = (w_0, w_1, \ldots, w_t, \ldots, w_n)$, we find a set of perturbation candidates $C$ using substitute\footnote{Substitute does not refer to word substitution here, but a `replacement' classifier. Specifically, one with an architecture and training data distinct from any target classifier $f$ we use.} classifier $f'$ to exhaustively generate new samples $D'$, any of which \emph{potentially} produces an incorrect label for a target classifier $f$. However, the samples are not selected based on such label changes, which therefore does not make this an adversarial attack. We follow and improve upon the adversarial substitution framework\footnote{\url{https://github.com/cmry/reap} \texttt{(ba8ee44)}} from \newcite{emmery-etal-2021-adversarial}, which in turn extends that of TextFooler \cite{DBLP:conf/aaai/JinJZS20}\footnote{\url{https://github.com/jind11/TextFooler}} with transformer-based perturbations.

\subsection{Selecting Words to Perturb} Target words $T(D,\ f')$ are selected and ranked based on their contribution to the classification of a document. This importance, or omission score \cite[among others]{DBLP:journals/tnn/SamekBMLM17,DBLP:journals/coling/KadarCA17} is calculated by deleting a word at a given position $D_t$, denoted as $D_{\setminus t}$. The omission score is then $o_y(D) - o_y(D_{\setminus t})$, where $o_y$ is the logit score of a substitute classifier $f'$. Intuitively, this would provide us with highly toxic words, or text parts related to bullying, which can be perturbed in some way.

\subsection{Proposing Perturbation Candidates} As we intend to improve lexical variation, we focus on proposing synonyms as perturbation candidates. \newcite{zhou-etal-2019-bert} condition BERT's masked language modeling on a given word by providing the original word its embedding to the masked position. They apply Dropout \cite{DBLP:journals/jmlr/SrivastavaHKSS14} as a surrogate mask, and show this to produce a top-$k$ of potential synonyms. The predicted words at the Dropout masked position by some separate transformer model $f_\text{cnd}$ are then our candidates $C(T, f_\text{cnd})$. To rank the candidates, they use a contextual similarity score:
\begin{equation}
\begin{split}
& \textsc{sim}\left(D, D^{\prime} ; t\right) =  \\ \sum_{i}^{n} \alpha_{i, t} \times
& \Lambda\left(\boldsymbol{h}\left(D_{i}\right), \boldsymbol{h}\left(D_{i}^{\prime} \right) \right)
\end{split}
\end{equation}
where: $\boldsymbol{h}\left(D_{i}\right)$ is the concatenation of $f_\text{cnd}$ its last four layers for a given $i^{th}$ token in document $D$, $D' = (w_0, \ldots, c_t, \ldots w_n)$ is the perturbed document $D$ where target word $w_t$ has been replaced with candidate $c$ at the index of $t$, $\Lambda$ is their cosine similarity, and $\alpha_{i, t}$ is the average self-attention score across all heads in all layers ranging from the $i^{th}$ token to the $t^{th}$ position in $D$. Finally, we sanitize the candidates: filtering single characters, plural and capitalized forms of the original words, sub-words, and sentence-level duplicates.

\subsection{Handling Out-of-vocabulary Words} BERT's associated tokenizers break down unknown words into word-pieces \cite{DBLP:journals/corr/WuSCLNMKCGMKSJL16}, meaning there is no single embedding to apply Dropout to. \newcite{zhou-etal-2019-bert} do not mention how they handle such cases; however, they are problematically common for our task (see Table~\ref{tab:lex-sub}). We therefore extend their method with a back-off method: if $w_t$ is out-of-vocabulary (OOV),\footnote{Note that the vocabulary of $f'$ might include tokens not contained in the vocabulary of BERT.} we collapse the word-pieces into one, and zero the embedding at that position (which then acts as a mask).\footnote{Alternative approaches, such as averaging and summing the token embeddings, did not provide better representations.} Words other than $w_t$ that are OOV remain word-pieces.

\section{Experimental Set-up}

We employ and compare this lexical substitution method to produce new positive instances. We evaluate if the perturbed documents hold up as adversarial samples, and if they can be used for data augmentation.

\subsection{Data} \label{subs:data}

\begin{table}[t!]
      \footnotesize
      \begin{tabular}{l|rrrrr}
      \toprule
                   & \includegraphics[width=9px]{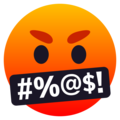}     & \includegraphics[width=9px]{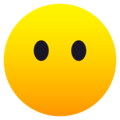}   & \textsc{ttr}   &  \textsc{avg tok/msg} \\  
      \midrule
        Ask.fm      & 5,001   & 89,404  & .154  & \ 12 \hfill ($\sigma = 23$) \ \ \\ 
        MySpace     & 426     & 1,627   & .016  & 391  \hfill ($\sigma = 285$) \\ 
        Twitter I   & 237     & 5,258   & .154  & \ 14 \hfill  ($\sigma = 8$) \ \ \ \ \\ 
        Twitter II  & 281     & 4,654   & .221  & \ 18 \hfill  ($\sigma = 8$)  \ \ \ \  \\ 
        YouTube     & 417     & 3,045   & .063  & 239  \hfill ($\sigma = 252$) \\ 
        \midrule
        Formspring  & 1,025   & 11,742  & .060  & \ 27 \hfill  ($\sigma = 29$) \ \ \\ 
        \bottomrule
    \end{tabular}
    \caption{Corpus statistics for cyberbullying data. Listed are the number of positive (\includegraphics[width=9px]{./gfx/angry.png}, bullying) and negative (\includegraphics[width=9px]{./gfx/nomouth.png}) instances, Type-Token Ratio (\textsc{ttr}), and the (rounded) average number of tokens per message (\textsc{avg tok/msg}), and their standard deviation ($\sigma$).
    }
    \label{tab:data-dsc}
\end{table}

For our corpora (all are English), we use two question-answering-style social networks that allow for anonymous posting: \emph{Formspring} \cite{DBLP:conf/icmla/ReynoldsKE11} and \emph{Ask.fm} \cite{DBLP:conf/ranlp/HeeLVMDPDH15}. The latter features multi-label annotation, but is binarized (any indication of bullying\footnote{Includes self-defenses and assistants of the victim.} is labeled positive) to be compatible with other corpora. These corpora are significantly larger than the rest, as their platforms are typically used by young adults, and notorious for their bullying content \cite{Binns2013}. Two long-form platforms can be found in \emph{YouTube} \cite{DBLP:conf/icwsm/DinakarRL11} and \emph{MySpace} \cite{bayzick2011detecting}, the latter of which has instances of ten posts. The smallest two are from \emph{Twitter}, both collected using topical keywords \cite{xu-etal-2012-learning,DBLP:conf/icis/BretschneiderWP14}. The corpora's statistics can be found in Table~\ref{tab:data-dsc}.

\subsection{Augmentation Models} \label{subs:aug-mod}

The models were implemented using HuggingFace's \texttt{transformers} \cite{wolf-etal-2020-transformers} library.\footnote{\url{https://huggingface.co/transformers/}} Dependency versions can be found in our repository.

\paragraph{Target Word Selectors} All experiments follow the same model-agnostic approach: target words are determined through substitute classifier $f'$ (i.e., a distinct model trained on a different corpus than used in any other experiments). Generally, this is Gaussian Naive Bayes over tf$\cdot$idf-weighted vectors, trained on \emph{Formspring}.  We additionally investigate a pre-trained version of BERT fine-tuned on the Jigsaw dataset \cite[\texttt{unitary/toxic-bert}]{Detoxify} as a transformer-based alternative for $f'$ (denoted by a +). While the task it has been fine-tuned on is slightly different, our assumption is that this model will have better representations and a larger vocabulary, which might make it more effective in choosing target words.

\paragraph{Substitutors}
We compare our implementation of \newcite{zhou-etal-2019-bert}'s substitutions (here: Dropout BERT) against other methods for masked word prediction: BERT \cite{DBLP:conf/naacl/DevlinCLT19} and BART \cite{lewis-etal-2020-bart}.  Table~\ref{tab:sub-comp} shows output examples of the various models. The probability of zeroing embedding dimensions in Dropout BERT is set to 0.2, as we empirically found that values around 0.3 often does not result in synonyms. We set the minimum required omission score to 0.005 for tokens to be considered for substitution, which yields 1-3 target words per document on average. The substitutions do not incorporate prior substitutions; they are done simultaneously---best candidates  first---and exhaustively (i.e., while candidates for all slots are available) for a maximum of five samples.

For all BERT models, we use the pre-trained \texttt{bert-\ large-cased}, for BART we use \texttt{bart-large}.\footnote{We empirically found the \texttt{bert-base} models to perform significantly worse at finding good synonyms.} We also report experiments where we use a fine-tuned toxicity version of pre-trained BERT \cite[\texttt{GroNLP/hateBERT}]{caselli-etal-2021-hatebert} for $f_{\text{cnd}}$ in Dropout BERT (here referred to as Hate BERT, and when using a fine-tuned substitute classifier: Hate BERT+). The idea here is similar to that of using Dropout BERT+; domain-specific vocabularies will likely result in better and more varied substitutions. Hate BERT(+) uses a different BERT-based toxicity model than the `+' model for $f'$, in order to keep these selections model-agnostic.

\paragraph{Baselines}

\begin{table}[t]
    \centering
    \footnotesize
    \begin{tabular}{l|p{5.7cm}}
        \toprule
        \textsc{prompt} & okay and stop calling me jaky you c\hlfancy{black}{o}ck \\
        \midrule
        \textsc{gpt-2} & s\hlfancy{black}{u}cker. you know it you f\hlfancy{black}{u}cking p\hlfancy{black}{u}ssy. you know you are an evil f\hlfancy{black}{u}cking b\hlfancy{black}{i}tch that only cares about getting her name in newspapers. i bet if you saw my face you wouldnt even believe i said ""oh yeah i think i can f\hlfancy{black}{u}ck you."" ... \\ [0.3em]
        \bottomrule
    \end{tabular}
    \caption{Output by GPT-2, receiving an original instance as prompt. The generated text (up to 70 tokens) is subsequently used as an augmented instance.}
    \label{tab:gpt-example}
\end{table}

The substitution models are compared with two baselines adapted from related work. Both have shown to improve toxicity detection \cite{gehman-etal-2020-realtoxicityprompts,quteineh-etal-2020-textual,DBLP:journals/corr/abs-2104-08826}, but have (to our knowledge) not been applied for augmenting cyberbullying content. Firstly, we employ the common data augmentation baseline: Easy Data Augmentation \cite[or EDA]{wei-zou-2019-eda}. EDA applies $n$ of the following operations to an input text: synonym replacement using WordNet \cite{DBLP:journals/cacm/Miller95}, random character insertions, swaps, and deletions. We set the number of augmentations made by EDA similar to that of our other models. Secondly, we employ fully unsupervised augmentation with GPT-2 \cite[implemented in the pre-trained \texttt{gpt2-large}]{radford2019language}. We use the positive instances (i.e., documents containing cyberbullying) of each dataset as prompt, with a maximum input length of 30 tokens, and the generated output length to a maximum of 70, as we found that toxicity is prevalent in the first part of the generation \cite[made similar observations]{gehman-etal-2020-realtoxicityprompts}. Table~\ref{tab:gpt-example} shows examples of the output, and the eventual divergence from toxicity.\footnote{GPT-2 in particular tends to descend into literary content after too many tokens are generated. We also experimented with GPT-3's \cite{DBLP:conf/nips/BrownMRSKDNSSAA20} \texttt{curie} from the OpenAI beta API (\url{https://beta.openai.com/}) but found systematically lower performance across all experiments compared to GPT-2. These results are therefore not included.}

\subsection{Classifiers} \label{subs:clf}
We follow recent state-of-the-art results \cite{DBLP:journals/access/ElsafouryKPR21} for our main classification model, and fine-tune all BERT-based models for 10 epochs with a batch size of 32 and a learning rate of $2e{-5}$, as suggested by \newcite{DBLP:conf/naacl/DevlinCLT19}. Accordingly, we set the maximum sequence length to 128, and insert a single linear layer after the pooled output. For the transformer experiments, we fine-tune incrementally: first on the original set, then on the augmented training set (including the original instances)---both using the same configuration (learning rate, batch size, etc.), except for running it for 2 epochs. This should offer performance advantages \cite{DBLP:journals/corr/abs-1904-06652}, as well as increase model stability.\footnote{We found that fine-tuning on the mixed set renders augmentation ineffective for all models we tested.} 

We compare BERT against a previously tried-and-tested \cite{DBLP:journals/corr/abs-1910-11922} `simple' linear baseline: the Scikit-learn \cite{DBLP:journals/jmlr/PedregosaVGMTGBPWDVPCBPD11} implementation of a Linear Support Vector Machine \cite[SVM ]{DBLP:journals/ml/CortesV95,DBLP:journals/jmlr/FanCHWL08} with binary Bag-of-Words (BoW) features, using hyperparameter ranges from \newcite{DBLP:journals/corr/abs-1801-05617}. Training of the SVM and BERT classifiers is done on a merged set of all the cyberbullying corpora in Table~\ref{tab:data-dsc}, except for \emph{Formspring} (reserved for substitute classifier $f'$)---always on the same 90\% split, augmented data or no. The SVM is tuned via grid search and nested, stratified cross-validation (with ten inner and three outer folds, no shuffling, using 10\% splits). The best settings (1-3-grams, class balancing, square hinge loss, and $C = 0.01$) are used in all experiments.

For both models, we also experiment with prepending a special token \cite[follow a similar approach]{daume-iii-2007-frustratingly,caswell-etal-2019-tagged} to the augmented instances (\texttt{<A>}). As per recommendations in \newcite{DBLP:journals/corr/abs-2003-02245}, the token is not added to the vocabulary. These models are referred to as either $f$ or $f_\text{aug}$ in Figure~\ref{fig:paper}, depending on if they were trained on augmented data. If not, we skip the 2 fine-tuning epochs for BERT.

\begin{table}[t!]
      \footnotesize
      \begin{tabular}{l|rr|rr}
      \toprule
                          & \multicolumn{2}{c}{\textsc{train}} &  \multicolumn{2}{c}{\textsc{test}} \\
                          \cmidrule{2-3} \cmidrule{4-5}
                          & \includegraphics[width=9px]{./gfx/angry.png}   & \includegraphics[width=9px]{./gfx/nomouth.png}  & \includegraphics[width=9px]{./gfx/angry.png}  &  \includegraphics[width=9px]{./gfx/nomouth.png} \\
      \midrule
        Merged            & 4,789         & 72,243        & 561           & 8,001   \\
        Augment Train     & 28,148        & 72,243        & 561           & 8,001   \\
        Augment Test      & 4,789         & 72,243        & 3,283        & 8,001       \\
      \bottomrule
    \end{tabular}
    \caption{Instance counts for the different splits used in our experiments. Augment Test is used for Experiment 1 (gauging the adversarial nature of our samples), Augment Train in Experiment 2 (data augmentation).}
    \label{tab:set-freq}
\end{table}

\subsection{Evaluation}

To evaluate our classifiers on the main classification task, we use $F_1$-scores. The impact of the substitution models on classification performance is measured via a decrease in True Positive Ratio (TPR) between regular and substituted samples (i.e., how many previously positively classified samples classified as negative after perturbation).  Note that in these experiments, $f'$ is the same (either Naive Bayes or BERT); therefore, the substitute classifier always chooses the same target words to perturb. The amount of samples depends on the quality of the candidates the models propose.

TPR decrease by itself might also indicate an augmented instance is not toxic anymore; hence, to evaluate the semantic consistency of the samples produced by the various augmentation models, we calculate both \textsc{Meteor} \cite{banerjee-lavie-2005-meteor,denkowski-lavie-2011-meteor} using the implementation from \texttt{nltk}\footnote{\url{https://www.nltk.org/_modules/nltk/translate/meteor_score.html} (v3.5)}, and \textsc{BERTScore} \cite{sellam-etal-2020-bleurt} between the original sentences and their respective augmented samples. \textsc{Meteor} measures flexible uni-gram token overlap, and \textsc{BERTScore} transformer-based similarity with respect to the contextual sentence encoding.

\subsection{Experiments}

We run our substitution pipeline (visualized in Figure~\ref{fig:paper}) on the positive instances $X_{\text{pos}}$ of some given corpus (or the entire collection), using the different models discussed in Section~\ref{subs:aug-mod} for $f'$ and $f_{\text{cnd}}$. Per such configuration, this generates augmented samples $X_{\text{pos}}'$ (up to five per original instance). These can either be classified as is, or mixed in with the original corpus, producing the augmented corpus $X'$, with $X_{\text{train}}'$, and $X_{\text{test}}'$ splits. Using this configuration, we run our three experiments:

\paragraph{Experiment 1} We gauge the \emph{lexical variation} (and hence the `adversarial' character) in our augmented samples via $f(X_{\text{pos}}')$. $F_1$-scores and TPR changes close to $f(X_{\text{pos}})$ imply the substitutions are similar to the original words. We confirm this meaning preservation through  semantic consistency metrics for $X_{\text{pos}}'$.

\paragraph{Experiment 2} Here, we train via the \emph{data augmentation} scheme discussed in Section~\ref{subs:clf}; i.e., fine-tune for 2 epochs on $f(X_\text{train}')$. The resulting augmented classifier is referred to as $f_\text{aug}$, which we evaluate on the original $X_{\text{test}}$. An increase in $F_1$-score with respect to $f(X_{\text{test}})$ indicates the augmentation is a success.

\paragraph{Experiment 3} We measure \emph{robustness} against perturbations, and \emph{transferability} via $f_{\text{aug}}(X_{\text{test}}')$ by evaluating  $f_{\text{aug}}$ performance across different substitution models producing perturbed samples in $X_{\text{test}}'$; i.e., in a many-to-many evaluation. Any TPR increase implies augmentation improves robustness against perturbations. A total TPR higher than $f(X_{\text{test}})$ (Plain) does not necessarily increase the $F_1$-score (from Experiment 2). If this increase holds for multiple perturbation models, this implies the augmentations are transferable.

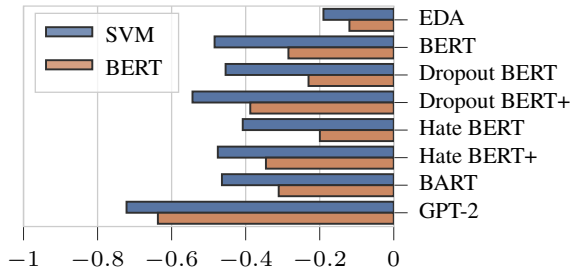
\begin{figure}
    \centering
\begin{tikzpicture}
\tikzstyle{every node}=[font=\scriptsize]
\definecolor{color0}{rgb}{0.347058823529412,0.458823529411765,0.641176470588235}
\definecolor{color1}{rgb}{0.798529411764706,0.536764705882353,0.389705882352941}

\begin{axis}[
axis line style={white!80!black},
legend cell align={left},
legend style={fill opacity=0.8, draw opacity=1, text opacity=1, draw=white!80!black},
legend pos=north west,
tick align=outside,
y=0.3cm,
y dir=reverse,
y grid style={white!80!black},
ymajorticks=true,
ymin=-0.5, ymax=7.5,
ytick style={color=white!15!black},
ytick={0,1,2,3,4,5,6,7},
ytick pos=upper,
yticklabels={EDA\phantom{p},BERT \phantom{p},Dropout BERT,Dropout BERT+,Hate BERT\phantom{p},Hate BERT+\phantom{p},BART\phantom{p},GPT-2\phantom{p}},
xtick pos=left,
x grid style={white!80!black},
xmajorgrids,
xmajorticks=true,
xmin=-1, xmax=0,
xtick style={color=white!15!black},
x=4cm
]
\addlegendimage{ybar,ybar legend,draw=white!18.8235294117647!black,fill=color0,semithick, legend image code/.code={\draw[] rectangle (0.5cm,0.1cm);}}
\addlegendentry{\ SVM}
\draw[draw=white!18.8235294117647!black,fill=color0,semithick] (axis cs:0,-0.4) rectangle (axis cs:-0.189968847352025,0);
\draw[draw=white!18.8235294117647!black,fill=color0,semithick] (axis cs:0,0.6) rectangle (axis cs:-0.483989722744181,1);
\draw[draw=white!18.8235294117647!black,fill=color0,semithick] (axis cs:0,1.6) rectangle (axis cs:-0.454219030520646,2);
\draw[draw=white!18.8235294117647!black,fill=color0,semithick] (axis cs:0,2.6) rectangle (axis cs:-0.543649579188982,3);
\draw[draw=white!18.8235294117647!black,fill=color0,semithick] (axis cs:0,3.6) rectangle (axis cs:-0.407928292829283,4);
\draw[draw=white!18.8235294117647!black,fill=color0,semithick] (axis cs:0,4.6) rectangle (axis cs:-0.47522303021216,5);
\draw[draw=white!18.8235294117647!black,fill=color0,semithick] (axis cs:0,5.6) rectangle (axis cs:-0.464097598760651,6);
\draw[draw=white!18.8235294117647!black,fill=color0,semithick] (axis cs:0,6.6) rectangle (axis cs:-0.721794392523365,7);

\addlegendimage{ybar,ybar legend,draw=white!18.8235294117647!black,fill=color1,semithick, legend image code/.code={\draw[] rectangle (0.5cm,0.1cm);}}
\addlegendentry{\ BERT}

\draw[draw=white!18.8235294117647!black,fill=color1,semithick] (axis cs:0,-2.77555756156289e-17) rectangle (axis cs:-0.119781931464174,0.4);
\draw[draw=white!18.8235294117647!black,fill=color1,semithick] (axis cs:0,1) rectangle (axis cs:-0.284541933504621,1.4);
\draw[draw=white!18.8235294117647!black,fill=color1,semithick] (axis cs:0,2) rectangle (axis cs:-0.230260896138126,2.4);
\draw[draw=white!18.8235294117647!black,fill=color1,semithick] (axis cs:0,3) rectangle (axis cs:-0.387605202754399,3.4);
\draw[draw=white!18.8235294117647!black,fill=color1,semithick] (axis cs:0,4) rectangle (axis cs:-0.19959495949595,4.4);
\draw[draw=white!18.8235294117647!black,fill=color1,semithick] (axis cs:0,5) rectangle (axis cs:-0.345265762051128,5.4);
\draw[draw=white!18.8235294117647!black,fill=color1,semithick] (axis cs:0,6) rectangle (axis cs:-0.310805577072037,6.4);
\draw[draw=white!18.8235294117647!black,fill=color1,semithick] (axis cs:0,7) rectangle (axis cs:-0.637121495327103,7.4);
\addplot [line width=1.08pt, white!26!black, forget plot]
table {%
0 -0.2
0 -0.2
};
\addplot [line width=1.08pt, white!26!black, forget plot]
table {%
0 0.8
0 0.8
};
\addplot [line width=1.08pt, white!26!black, forget plot]
table {%
0 1.8
0 1.8
};
\addplot [line width=1.08pt, white!26!black, forget plot]
table {%
0 2.8
0 2.8
};
\addplot [line width=1.08pt, white!26!black, forget plot]
table {%
0 3.8
0 3.8
};
\addplot [line width=1.08pt, white!26!black, forget plot]
table {%
0 4.8
0 4.8
};
\addplot [line width=1.08pt, white!26!black, forget plot]
table {%
0 5.8
0 5.8
};
\addplot [line width=1.08pt, white!26!black, forget plot]
table {%
0 6.8
0 6.8
};
\addplot [line width=1.08pt, white!26!black, forget plot]
table {%
0 0.2
0 0.2
};
\addplot [line width=1.08pt, white!26!black, forget plot]
table {%
0 1.2
0 1.2
};
\addplot [line width=1.08pt, white!26!black, forget plot]
table {%
0 2.2
0 2.2
};
\addplot [line width=1.08pt, white!26!black, forget plot]
table {%
0 3.2
0 3.2
};
\addplot [line width=1.08pt, white!26!black, forget plot]
table {%
0 4.2
0 4.2
};
\addplot [line width=1.08pt, white!26!black, forget plot]
table {%
0 5.2
0 5.2
};
\addplot [line width=1.08pt, white!26!black, forget plot]
table {%
0 6.2
0 6.2
};
\addplot [line width=1.08pt, white!26!black, forget plot]
table {%
0 7.2
0 7.2
};
\end{axis}

\end{tikzpicture}
    \caption{Decrease in True Positive Rate of the SVM and BERT classifiers after the respective substitution models have been applied (lower is more adversarial).}
    \label{fig:tpr-dec}
\end{figure}

\section{Results and Discussion}

Here, we discuss the results of our three Experiments (Sections~\ref{subs:exp1}-\ref{subs:exp3}) and close with suggestions for future work. The main results can be found in Table~\ref{tab:exp1-2} and \ref{tab:exp3}.

\subsection{The Effect of Lexical Variation} \label{subs:exp1}

The results for this experiment can be found under the `Samples' row in Table~\ref{tab:exp1-2}.

\paragraph{Classifier Performance} It can be seen that unsupervised (prompt conditioning) samples (i.e., GPT-2) are the most difficult to classify. This is to be expected, as the generated output is not always toxic. However, it is arguably rather remarkable that a large amount of the generated sentences are labeled positive by the cyberbullying classifier. This confirms that (contextually more) harmful content is generated (as illustrated in Table~\ref{tab:gpt-example}), as also shown for toxicity detection by \newcite{gehman-etal-2020-realtoxicityprompts} and \newcite{DBLP:conf/acl/OusidhoumZFSY20}. Moving on, we can see the fine-tuned target selector models (`+') show most `adversarial' behavior, likely providing lexical diversity on words more important to the content classification. BART induces similar performance drops, but often inserts noise (see Table~\ref{tab:sub-comp}). The Dropout models seem to produce samples that are less diverse, but still show a solid .1 drop in $F_1$-score (17.67\% on average). To emphasize, this decrease is based on \emph{untargeted} substitutions; i.e., without selecting the substituted words as to change the predictions of either $f$ or $f'$.

\paragraph{Adversarial Samples}

\begin{figure}
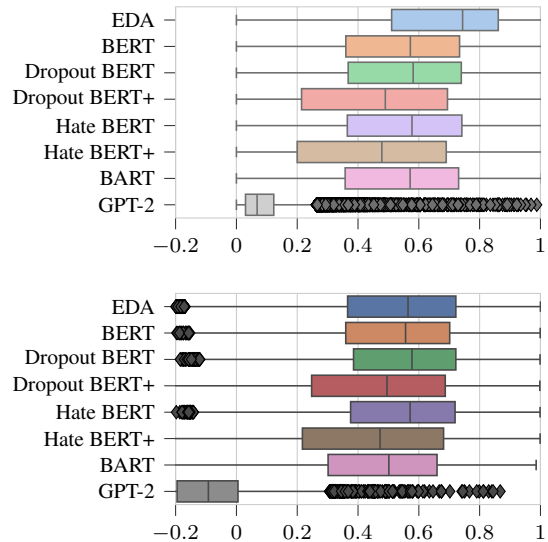

    \centering
    \includestandalone{./gfx/boxplot}
    \caption{Semantic consistency metrics (higher is better): \textsc{Meteor} (upper) and \textsc{BERTScore} (lower) scores per model---evaluating the augmented samples (positive only) with the original documents as a reference.}
    \label{fig:sem-con}
\end{figure}

Additional analyses can be found in Figure~\ref{fig:tpr-dec}. We observe the same patterns per substitution model\footnote{To equalize the length, we matched the amount of non-augmented test set instances for this experiment.} as in Table~\ref{tab:exp1-2}, with the BERT classifier showing to suffer around 20\% less in TPR compared to the SVM. This difference can partly be explained by the substitution models sampling from the same model as we fine-tuned for the classification task (\texttt{bert-large-cased}). As can be observed in this Figure, the difference is smaller when this is not the case (`+' models). This experiment not only underlines the strong focus on lexical cues\footnote{Which raises its own issues; see e.g., and \newcite{zhou-etal-2021-challenges}, for work on bias and debiasing.} from linear classifiers, but also that transformer models are not immune to lexical variation---even when candidates are sampled from their own language model. This provides further evidence in line with research from \newcite{DBLP:journals/access/ElsafouryKPR21}, and \newcite{DBLP:conf/sigir/ElsafouryKWR21} (see Section~\ref{sec:rel-work}).

\begin{table*}[t!]
    \centering
    \footnotesize
    \begin{tabular}{l|>{\raggedleft\arraybackslash}p{1.1cm}||>{\raggedleft\arraybackslash}p{1.1cm}|>{\raggedleft\arraybackslash}p{1.1cm}|>{\raggedleft\arraybackslash}p{1.1cm}|>{\raggedleft\arraybackslash}p{1.1cm}|>{\raggedleft\arraybackslash}p{1.1cm}|>{\raggedleft\arraybackslash}p{1.1cm}|>{\raggedleft\arraybackslash}p{1.1cm}|>{\raggedleft\arraybackslash}p{1.1cm}}
      \toprule
                    &                                               &                                 &                          & Dropout                          & Dropout                 & Hate                              & Hate &        &  \\
        $X_{\text{train}}$  &    Plain                              & EDA                             & BERT                     &  BERT                            & BERT+                   &  BERT                             & BERT+                   & BART                   & GPT-2 \\
      \midrule 
                    &                                               & \multicolumn{8}{c}{$f(X_{\text{pos}}')$} \\
      \midrule
        Merged &                           .614 \scriptsize{.009}   &         .598 \scriptsize{.012}  & .458 \scriptsize{.002}  & .491 \scriptsize{.005}            & .439 \scriptsize{.002}  &         .520 \scriptsize{.005}    & .478 \scriptsize{.004}  & .436 \scriptsize{.007} &          .334 \scriptsize{.007} \\
      \midrule \midrule
              & &    \multicolumn{8}{c}{$f_{\text{aug}}(X_{\text{test}})$} \\
      \midrule
        Merged      & \cellcolor{PaleRed} .563 \scriptsize{.014}    &         .553  \scriptsize{.007} & .538 \scriptsize{.014}  & .546 \scriptsize{.012}            & .523 \scriptsize{.004}  & \textbf{.562} \scriptsize{.007}   & .535 \scriptsize{.009}  & .536 \scriptsize{.013} &          .550 \scriptsize{.017} \\ 
      \midrule
        Ask.fm      & \cellcolor{PaleRed} .621 \scriptsize{.011}    &         .591  \scriptsize{.011} & .581 \scriptsize{.016}  & .597 \scriptsize{.015}            & .574 \scriptsize{.003}  & \textbf{.611} \scriptsize{.007}   & .592 \scriptsize{.007}  & .587 \scriptsize{.015} &          .601 \scriptsize{.009}\\ 
        Myspace     &                       .436 \scriptsize{.093}  &         .476  \scriptsize{.021} & .403 \scriptsize{.058}  & .376 \scriptsize{.161}            & .351 \scriptsize{.040}  &         .414  \scriptsize{.042}   & .370 \scriptsize{.065}  & .387 \scriptsize{.043} & \textbf{.496} \scriptsize{.037} \\
        Twitter I   &                       .596 \scriptsize{.046}  & \textbf{.630} \scriptsize{.016} & .592 \scriptsize{.059}  & .531 \scriptsize{.048}            & .594 \scriptsize{.030}  &         .617  \scriptsize{.053}   & .533 \scriptsize{.051}  & .591 \scriptsize{.027} &          .583 \scriptsize{.097} \\
        Twitter II  &                       .308 \scriptsize{.059}  &         .290  \scriptsize{.038} & .297 \scriptsize{.043}  & \textbf{.329} \scriptsize{.034}   & .257 \scriptsize{.040}  &         .313  \scriptsize{.048}   & .268 \scriptsize{.027}  & .277 \scriptsize{.077} &          .295 \scriptsize{.048} \\ 
        YouTube     &                       .150 \scriptsize{.033}  & \textbf{.226} \scriptsize{.057} & .180 \scriptsize{.010}  & .201 \scriptsize{.066}            & .144 \scriptsize{.045}  &         .173  \scriptsize{.056}   & .152 \scriptsize{.009}  & .152 \scriptsize{.055} &          .207 \scriptsize{.061} \\ 
      \bottomrule
    \end{tabular}
    \caption{BERT-based cyberbullying classification scores ($F_1$) for Experiments 1 (under $f(X_{\text{pos}}')$) and 2 (under $f_{\text{aug}}(X_{\text{test}})$). Classifiers are trained and tested on the indicated corpus (from Section~\ref{subs:data}), `Merged` is their combination. The other columns indicate, respectively: no substitutions (Plain), EDA, BERT-based models (where `+' indicates $f'$ uses BERT rather than an SVM, and `Hate' that $f_{\text{cnd}}$ is pre-trained), BART, and GPT-2. Highlighted cells indicate that non-augmented performance was highest, bold indicates the highest performance per augmentation model. Standard deviation (small script) is reported over five runs with different seeds.}
    \label{tab:exp1-2}
\end{table*}

\paragraph{Semantic Consistency of Samples}

Here, we compared $X'_{\text{pos}}$ with $X_{\text{pos}}$ as a reference. The results for \textsc{Meteor} and \textsc{BERTScore} of these pairs can be found in Figure~\ref{fig:sem-con}. Generally, these confirm the trend from the previous two parts of the experiment: models that have higher semantic consistency have less effect on classification performance. A clear difference in \textsc{Meteor} can be observed between EDA and the other models. This is likely due to both the metric and model using WordNet, resulting in bias in favor of EDA. GPT-2 is a strong outlier, as it generates new data. 

The semantic consistency scores seem comparable, and at times slightly better, than previous lexical substitution work \cite[although these are all explicitly adversarial]{DBLP:conf/uss/ShettySF18,DBLP:conf/coling/EmmeryAC18,DBLP:journals/corr/abs-2011-03901}. We noticed that regarding the samples themselves, the transformer-based models often noticeably break down in terms of semantic preservation for the lower ranked candidates (see Table~\ref{tab:sub-comp}). For the models that do not use soft semantic constraints (such as Dropout), we already find antonyms, and generally ungrammatical and incoherent sentences within the top 5 candidates. Interestingly, at the same time, \textsc{BERTScore} assigns comparable scores to antonyms as it does for (intuitively) better substitution candidates. Given these observations, if mimicking adversarial behavior is to be given more weight, one should consider limiting the number of augmented samples, and tuning the omission score cut-off might prove to be worthwhile.

\begin{table}
    \setlength{\tabcolsep}{3.5pt}
    \centering
    \footnotesize
    \begin{tabular}{l|l|llll>{\columncolor{AquaBlue}}l>{\columncolor{AquaBlue}}lll|l}
        \toprule
        &                \# &            &   &       &     &  \multicolumn{2}{c}{\textsc{targets}}   & & & \textsc{bsc}  \\
                         \cmidrule{7-8}
                            &            & I & don't & get & why & \pale{people} & \pale{like} & u & .    &      \\
        \midrule
        \multirow{2}{*}{BE} & \textsc{1} & I & don't & get & why & they & want & u & .      & .809 \\
                            & \textsc{5} & I & don't & get & why & boys & hate & u & .      & .807 \\
        \midrule
        Dr                  & \textsc{1} & I & don't & get & why & everyone & want & u & .  & .862 \\
        BE                  & \textsc{5} & I & don't & get & why & you & love & u & .       & .806 \\
        \midrule
        \multirow{2}{*}{BA} & \textsc{1} & I & don't & get & why & you & are & u & .        & .700 \\
                            & \textsc{5} & I & don't & get & why & so & have & u & .        & .634 \\
        \bottomrule
    \end{tabular}
    \caption{Augmentations including the top 1 and 5 candidates from BERT (BE), Dropout BERT (Dr BE), and BART (BA), and the \textsc{BERTScore} (\textsc{bsc}) using the original text as reference, showing quality degradation (not well reflected in the metric) when sample size increases. BERT suggests antonyms, BART fails semantically.}
    \label{tab:sub-comp}
\end{table}

\subsection{Substitutions for Data Augmentation}

\begin{table*}
\footnotesize
\setlength{\tabcolsep}{4.2pt}
\centering
\begin{tabular}{r|>{\raggedleft\arraybackslash}p{1.1cm}||>{\raggedleft\arraybackslash}p{1.1cm}|>{\raggedleft\arraybackslash}p{1.1cm}|>{\raggedleft\arraybackslash}p{1.1cm}|>{\raggedleft\arraybackslash}p{1.1cm}|>{\raggedleft\arraybackslash}p{1.1cm}|>{\raggedleft\arraybackslash}p{1.1cm}|>{\raggedleft\arraybackslash}p{1.1cm}|>{\raggedleft\arraybackslash}p{1.1cm}}
\toprule
              & \textsc{initial} & \multicolumn{8}{c}{\multirow{2}{*}{$f_{\text{aug}}$}} \\
              & \textsc{tpr}      \\
              \midrule
              &                     &                               &                               &         Dropout               &         Dropout                   &      Hate                     &      Hate                     &                               &        \\
Plain         &   .537              & EDA                           & BERT                          &         BERT                  &         BERT+                     &      BERT                     &      BERT+                    & BART                          & GPT-2  \\
\midrule
$X_{\text{test}}' \downarrow$    &  &  \multicolumn{8}{c}{\textsc{$\Delta$ tpr}} \\
\midrule
EDA           & .498                &  \cellcolor{PalePurple} .270  &  .106                         &    .104                       &     .108                          &  .101                         &  .107                         &  \textbf{.114}                &  .037 \\
BERT          & .390                & -.033                         &  \cellcolor{PalePurple} .195  &    .144                       &     .110                          &  .128                         &  .092                         &  \textbf{.149}                &  .085 \\
Dropout BERT  & .421                & -.017                         &  \textbf{.183}                & \cellcolor{PalePurple} .183   &     .143                          &  .143                         &  .111                         &  .152                         &  .086 \\
Dropout BERT+ & .362                &  .015                         &  .228                         &    .236                       &     \cellcolor{PalePurple} .301   &  .177                         &  \textbf{.238}                &  .191                         &  .088 \\
Hate BERT     & .444                & -.020                         &  \textbf{.170}                &    .148                       &     .104                          &  .162                         &  .123                         &  .143                         &  .073 \\
Hate BERT+    & .394                &  .034                         &  .188                         &    .174                       &     \textbf{.238}                 &  .215                         &  \cellcolor{PalePurple} .262  &  .183                         &  .065 \\
BART          & .378                & -.010                         &  \textbf{.200}                &    .157                       &     .127                          &  .142                         &  .115                         &  \cellcolor{PalePurple} .243  &  .079 \\
GPT-2         & .303                & -.098                         &  .031                         &    .003                       &       -.020                       &  .003                         & -.025                         &  \textbf{.048}                &  \cellcolor{PalePurple} .597 \\
\midrule
\textsc{mean}  & .399                & -.114 &  \textbf{.158} &    .138 &     .116 &  .111 &  .130 &  .140 &  .073 \\
\bottomrule
\end{tabular}
\caption{Transferability and robustness of various $f_\text{aug}$ models on various augmented test samples $X_{\text{test}}'$. Shown is the original True Positive Rate (TPR, under \textsc{initial trp}) for $f$ (Plain), and the change ($\Delta$) in TPR of $f_{\text{aug}}(X_{\text{test}}')$ with respect to $f(X_{\text{test}}')$. Positive $\Delta$ \textsc{trp} shows robustness to perturbations after augmentation, negative the opposite. The classifiers most robust against same-model perturbations are highlighted, in bold the next-best. At the bottom, \textsc{mean} (per augmented classifier) TPR is shown, \emph{excluding} performance on the same model (to reflect transferability).}
\label{tab:exp3}
\end{table*}

Given the strong performance effect the substitution models had on our classifiers, it seems plausible that they might prove to produce effective samples for augmentation purposes. Looking at the lower portion of Table~\ref{tab:exp1-2}; however, we can see that augmentation does not improve performance on the two biggest sets (Merged, and Ask.fm). Dropout BERT seems to improve performance for one of the smaller sets (Twitter II), but overall, EDA is generally a close contender with, if not more effective, than all of the more `advanced' models. Interestingly, the transformer-based models seem to yield sizable improvements on the Myspace set, with GPT-2 increasing it most. The latter might be attributed to the low Type-Token Ratio in this set (see Table~\ref{tab:data-dsc}).

\paragraph{Interpretation} Generally, none of these methods (baselines, or substitution-based augmentation) seem to yield the same performance improvement as observed in toxicity work \cite{DBLP:conf/icmla/IbrahimTE18,Jungiewicz_Smywinski-Pohl_2019}. However, note that, as we were interested in simulating potentially adversarial behavior, we conducted model-agnostic augmentation (that is, given an unknown attacker, or noise). Hence, while we might employ these models in an explicit adversarial training scheme to directly improve model performance, this would require extensive transferability evaluations---typically requiring larger, higher quality datasets---and only satisfy one dimension (data). Given this, we argue an improvement in classifying the augmented sets, as in Experiment 1 (Section~\ref{subs:exp1}), is more significant.

\subsection{Augmentation for Robustness} \label{subs:exp3}

The results of Experiment 3 can be found in Table~\ref{tab:exp3}. We report TPR changes for $f(X_{\text{test}})$ (under initial TPR), and  $f_{\text{aug}}(X_{\text{test}}')$ per setting to create $f_{\text{aug}}$ and $X'$ respectively.

\paragraph{Robustness}

Generally, it can be observed that (unsurprisingly) the augmented classifiers increase TPR most when the substitutions come from the same type of model. Hence, the `second-best' TPR increases are more interesting. It can be observed that BERT and BART show strongest TPR improvements on three sets respectively, followed by the `+' models with one set respectively. This is quite a remarkable, contrasting result to Experiments 1 and 2, although it aligns with the observations from the semantic consistency scores that more conservative models are less effective augmenters. Hence, it seems that substitutions that are \emph{more} diverse, and distant from the original instances provide better robustness against perturbations. While their output might be less semantically consistent, this is generally not a relevant criterion when one is only interested in improving task robustness.

\paragraph{Transferability} Systematic performance gain across all substitution models (i.e., transferability) is the final indicator of augmentation utility. First, it must be noted that the TPR differences between `same-model' and distinct model pairs are smaller for the transformer-based models (.026 on average) than EDA (.156). Lexical substitution using out-of-the-box BERT---in addition to high robustness---also achieves the highest transferability (mean .158) across substituted sets.

\paragraph{Performance Trade-Off}

The $f_{\text{aug}}$ results from this experiment should be contextualized against the performance trade-off in $F_1$-score from Experiment 2. Using the information in Table~\ref{tab:exp3}, it can be inferred that the best performing model, BERT, actually improves absolute TPR on average; if we add its .158 mean TPR increase to the .399  average (= .557) this exceeds the .537 non-augmented TPR. However, as we showed in Experiment 2 (Table~\ref{tab:exp1-2}), this does not improve overall task performance; rather, it decreases performance. For BERT, the $F_1$-score slightly drops (.025-.030) on all sets. Hence, this is not a silver bullet, and such trade-offs should be considered when deploying these augmentation models to improve robustness against lexical variation.

\paragraph{Limitations and Future Work}

A substantial hurdle toward deploying the presented models for augmentation purposes is time. Upsampling the positive instances shown in Table~\ref{tab:set-freq} (5,350 total) with the transformer-based models takes 2-3 hours per model on a single NVIDIA Titan X (Pascal).\footnote{Training the BERT classifiers takes up to 31 hours, augmentation 40 minutes, predictions on the test set 5 minutes.} This impacts the amount of parameters that can be tweaked in reasonable time when using this architecture (such as omission score cut-offs, cosine similarity when ranking, dropout values, etc. which we all set empirically). Such computational demand is acceptable for smaller datasets like ours, and the augmentations can be run `offline' (i.e., one time only), but these limitations should certainly be taken into account when scaling is among one's desiderata.

Hence, recent work on decreasing the amount of queries for related models \cite{DBLP:journals/corr/abs-2106-07047} is particularly relevant for future work. Additionally, there is a myriad of components the base architecture we presented here could be improved with. Most are discussed in \newcite{emmery-etal-2021-adversarial}; however, some new work is specifically of interest to data augmentation, such as improving the substitutions using beam search \cite[as opposed to the simultaneous rollout we used in the current work]{DBLP:journals/corr/abs-2110-08036}. More broadly, adversarial training \cite{DBLP:conf/acl/SiZQLWLS21,DBLP:journals/corr/abs-2107-10137}, implementing more robust stylometric features \cite{markov-etal-2021-exploring}, or model-based weightings of the augmentation models could be explored; e.g., by selecting instances with a generation model in the loop \cite{DBLP:conf/aaai/Anaby-TavorCGKK20}. This could be a particularly worthwhile option when focusing on conversation scopes, rather than message-level cyberbullying content \cite{DBLP:journals/corr/abs-1910-11922}.

\section{Related Work} \label{sec:rel-work}

Our work combines multiple sizeable---to the extent that they respectively produced several surveys \cite{DBLP:journals/csur/FortunaN18,gunasekara-nejadgholi-2018-review,DBLP:journals/corr/abs-1908-06024,banko-etal-2020-unified,madukwe-etal-2020-data,DBLP:journals/fi/MuneerF20,DBLP:journals/taffco/SalawuHL20,DBLP:journals/corr/abs-2106-00742,DBLP:journals/csur/MladenovicOS21}---areas of research; hence, we will provide a concise overview of the work directly related to our experimental setup.

For all tasks, the issue of generalization seems a particularly popular subject of study: for cyberbullying, \newcite{DBLP:journals/corr/abs-1910-11922}, and \newcite{DBLP:conf/asunam/LarochelleK20}, conclude there is little consensus in labeling practices, overlap between datasets, and that a combination of all datasets seems to transfer performance best. For hate speech, \newcite{DBLP:journals/hcis/SalminenHCJAJ20}, and \newcite{DBLP:journals/ipm/FortunaCW21}, draw similar conclusions, showing that general forms of harm (e.g., toxic, offensive) generalize better than specific ones, such as hate speech. Finally, \newcite{nejadgholi-kiritchenko-2020-cross} provide unsupervised suggestions to address topic bias in data curation, potentially improving generalization. We draw from these works through cross-domain experiments on individual and combined corpora for cyberbullying, as well as pre-training on more general subtasks such as toxicity. 

Recent cyberbullying work \cite[e.g., are seminal work]{DBLP:conf/icmla/ReynoldsKE11,xu-etal-2012-learning,nitta-etal-2013-detecting,DBLP:conf/icis/BretschneiderWP14,DBLP:conf/ai/DadvarTJ14,van-hee-etal-2015-detection} has primarily focused on deploying Transformer-based models \cite{DBLP:conf/nips/VaswaniSPUJGKP17}; by and large fine-tuning \cite[e.g.]{swamy-etal-2019-studying,paul_cyberbert_2020,DBLP:journals/internet/Gencoglu21}, or re-training \cite{DBLP:journals/corr/abs-2010-12472} BERT. It is worth noting that \newcite{DBLP:journals/access/ElsafouryKPR21,DBLP:conf/sigir/ElsafouryKWR21} show that although fine-tuning BERT achieves state-of-the-art performance in classification, its attention scores do not correlate with cyberbullying features, and they expect generalization of such models to be subpar. In our experiments, we employ similar domain-specific fine-tuned BERT models, and gauge generalization, sensitivity to perturbations, and the effects of augmentation to potentially improve the former.

Adversarial attacks on text \cite[e.g., provide broader surveys]{DBLP:journals/tist/ZhangSAL20,DBLP:journals/corr/abs-2103-00676} can roughly be divided in character-level and word-level. The former relates to purposefully misspelling or otherwise symbolically replacing text (e.g., \emph{fvk you}, \emph{@ssh*l3}) to subvert algorithms \cite{eger-etal-2019-text,DBLP:journals/corr/abs-1912-06872}. \newcite{DBLP:conf/acl-alw/WuKS18} show such attacks on toxic content can be effectively deciphered. Word-level attacks are arguably straight-forward for humans, but significantly more challenging to automate---requiring preservation of toxicity; i.e, the semantics of the sentence. Previous work has investigated the effect of minimal edits on high-impact toxicity words, replacing them with harmless variants \cite{DBLP:journals/corr/HosseiniKZP17,brassard-gourdeau-khoury-2019-subversive}. Our current work is similar to that of \newcite{DBLP:conf/micai/Tapia-TellezE20}, and closest to that of \newcite{DBLP:conf/sepln/Guzman-Silverio20}, who apply simple synonym replacement using EDA, as well as adversarial token substitutions---the latter using TextFooler  on misclassified instances. We extend BERT-based lexical substitution \cite{zhou-etal-2019-bert} for model-agnostic perturbations, and data augmentation.

Finally, regarding data augmentation for online harms \cite[among others, provide more general-purpose overviews for various natural language data]{DBLP:journals/corr/abs-2107-03158,feng-etal-2021-survey}, toxicity work partly overlaps with work on adversarial attacks on text; for example, the synonym replacement from \newcite{DBLP:conf/icmla/IbrahimTE18}, and \newcite{Jungiewicz_Smywinski-Pohl_2019}, which are distinctly either unsupervised, or semi-supervised. Another such example can be found in \newcite{rosenthal-etal-2021-solid} employed democratic co-training to collect a large corpus of toxic tweets, and \newcite{gehman-etal-2020-realtoxicityprompts} find triggers that produce toxic content, querying GPT-like models \cite{radford2018improving}. Fully unsupervised augmentation has also been employed in \newcite{quteineh-etal-2020-textual}, and \newcite{DBLP:journals/corr/abs-2104-08826}. In our experiments, we use a pipeline of models for lexical substitution, and compare it to GPT generations \cite{radford2019language,DBLP:conf/nips/BrownMRSKDNSSAA20}.

\section{Conclusion}

In this work, we employed model-agnostic, transformer-based lexical substitutions to the task of cyberbullying classification. We show these perturbations significantly decrease classifier performance. Augmenting them using perturbed instances as new samples slightly trades off task performance with improved robustness against lexical variation. Future work should further investigate the use of these models to simulate and mitigate the effect of adversarial behavior in content moderation. 

\section{Acknowledgments}

Our research strongly relied on openly available resources.  We thank all whose work we could use. 

\section{Bibliographical References}\label{reference}

\bibliographystyle{lrec2022-bib}
\bibliography{the-anthology,refs}


\end{document}